\documentclass[nohyperref]{article}

\pdfoutput=1

\usepackage{microtype}
\usepackage{graphicx}
\usepackage{subfigure}
\usepackage{booktabs} 
\usepackage{svg}

\usepackage{hyperref}
\usepackage{flushend}

\usepackage[accepted]{icml2022}

\usepackage{amsmath}
\usepackage{amssymb}
\usepackage{mathtools}
\usepackage{amsthm}
\usepackage{enumitem}
\usepackage{bbding}
\usepackage{multicol}
\usepackage{multirow}
\usepackage{makecell}

\usepackage[capitalize,noabbrev]{cleveref}

\theoremstyle{plain}

\theoremstyle{definition}

\theoremstyle{remark}

\usepackage[textsize=tiny]{todonotes}

\definecolor{MyDarkBlue}{rgb}{0,0.08,1}
\definecolor{MyDarkGreen}{rgb}{0.02,0.6,0.02}
\definecolor{MyDarkRed}{rgb}{0.8,0.02,0.02}
\definecolor{MyDarkOrange}{rgb}{0.40,0.2,0.02}
\definecolor{MyPurple}{RGB}{111,0,255}
\definecolor{MyRed}{rgb}{1.0,0.0,0.0}
\definecolor{MyGold}{rgb}{0.75,0.6,0.12}
\definecolor{MyDarkgray}{rgb}{0.66, 0.66, 0.66}

\newcommand{\myparagraph}[1]{\vspace{-5pt}\paragraph{#1}}
\newcommand{\myitem}[1]{\vspace{-4pt}\item{#1}}
\newcommand{\model}{NeuroFluid}
\newcommand{\render}{PhysNeRF}

\icmltitlerunning{NeuroFluid: Fluid Dynamics Grounding with Particle-Driven Neural Radiance Fields}

\begin{document}

\twocolumn[
\icmltitle{NeuroFluid: Fluid Dynamics Grounding with Particle-Driven \\Neural Radiance Fields}

\begin{icmlauthorlist}
\icmlauthor{Shanyan Guan}{sch}
\icmlauthor{Huayu Deng}{sch}
\icmlauthor{Yunbo Wang}{sch}
\icmlauthor{Xiaokang Yang}{sch}
\end{icmlauthorlist}

\icmlaffiliation{sch}{MoE Key Lab of Artificial Intelligence, AI Institute, Shanghai Jiao Tong University, Shanghai 200240, China}

\icmlcorrespondingauthor{Yunbo Wang}{yunbow@sjtu.edu.cn}

\icmlkeywords{Machine Learning, ICML}

\vskip 0.3in

]

\printAffiliationsAndNotice{}  

\begin{abstract}

Deep learning has shown great potential for modeling the physical dynamics of complex particle systems such as fluids. Existing approaches, however, require the supervision of consecutive particle properties, including positions and velocities. In this paper, we consider a partially observable scenario known as \textit{fluid dynamics grounding}, that is, inferring the state transitions and interactions within the fluid particle systems from sequential visual observations of the fluid surface. We propose a differentiable two-stage network named \textit{NeuroFluid}. Our approach consists of (i) a particle-driven neural renderer, which involves fluid physical properties into the volume rendering function, and (ii) a particle transition model optimized to reduce the differences between the rendered and the observed images. NeuroFluid provides the first solution to unsupervised learning of particle-based fluid dynamics by training these two models jointly. It is shown to reasonably estimate the underlying physics of fluids with different initial shapes, viscosity, and densities. 
\end{abstract}

\section{Introduction}

\begin{figure}[t]
\vskip 0.2in
\begin{center}
\centerline{\includegraphics[width=\columnwidth]{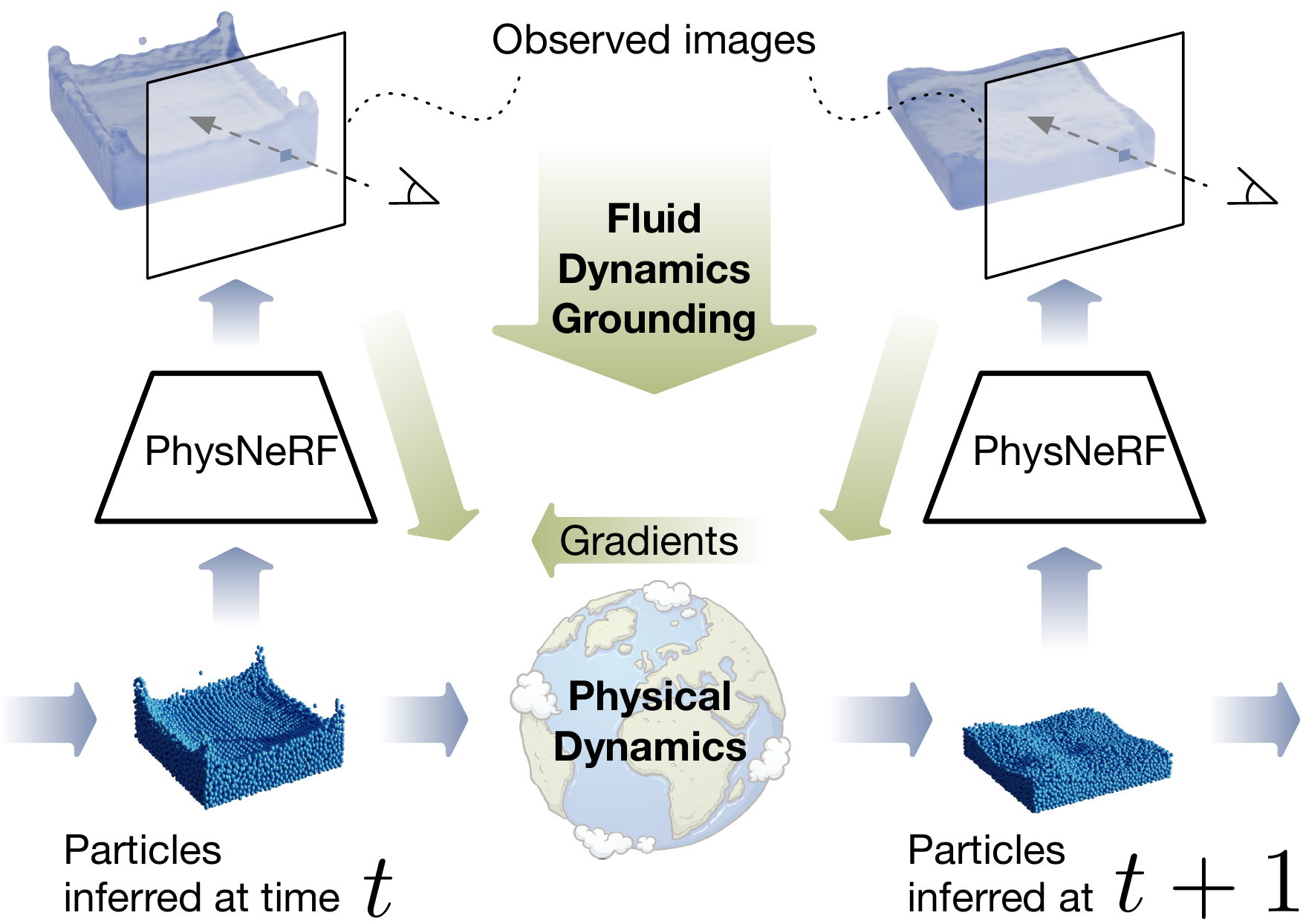}}
\caption{Fluid dynamics grounding is to reason about the underlying physical dynamics in fluid particle systems from sequential visual observations. We propose \textit{\model{}}, the first fully differentiable solution to this problem, in which a key component is the particle-driven neural renderer named \textit{\render{}}.}
\label{fig:intro}
\end{center}
\vskip -0.2in
\end{figure}

Intuitive physics or intuitive physical inference is a research area that gathers a lot of interest in the machine learning community.
Recent methods mainly focus on building neural models to reason about stability, collisions, forces, and velocities from images or videos~\cite{battaglia2013simulation,wu2017learning}.
In this paper, we explore a new problem in intuitive physics, namely \textit{fluid dynamics grounding}, defined as inferring the physical dynamics of fluids from a sequence of 2D visual observations collected from a sparse set of views.
As shown in Figure \ref{fig:intro}, to understand the underlying physics, the key is to solve the inverse problem of synthesizing the visual scenes. 
A typical forward modeling process includes two steps: fluid simulation and dynamic scene rendering.

Despite recent progress in learning particle-based fluid simulators~\cite{li2018learning,ummenhofer2019lagrangian,kim2019deep,sanchez2020learning}, it remains an open question \textit{whether neural networks can infer fluid dynamics directly from observed images}, since all existing work requires engineered simulators to provide consecutive particle locations as training data.
To answer this question, we propose \textit{\model{}}, the first fully differentiable solution to fluid dynamics grounding. 
The key idea is to link particle-based fluid simulation with particle-driven neural rendering in an end-to-end trainable framework, such that the two networks can be jointly optimized to obtain reasonable particle representations between them.

However, since we generally do not have any prior knowledge of the physical properties of the fluid in this intuitive physics setup, it is difficult to impose learning constraints on the particle representations directly.
Therefore, to keep fluid dynamics grounding from an undesirable trivial solution to ``de-rendering'' the visual observations, we propose \textit{\render{}}, a neural renderer in forms of a particle-driven and geometry-dependent \textit{neural radiance field} (NeRF). 
It takes as input the estimated particles and calculates their geometric relations with the na\"ive sampling points of NeRF that emit view-dependent radiance.
We use \render{} as a key component of \model{} to allow grounding non-trivial physical dynamics from visual observations.

\model{} is evaluated on dynamic scenes of fluids with different initial shapes, viscosity, and densities.
It achieves competitive results in the accuracy of particle state inference and the quality of novel view synthesis and future rollouts.

The contributions of this paper are summarized as follows:
\vspace{-4pt}
\begin{itemize}[leftmargin=*]
    \myitem We present and explore a new research field in intuitive physics, which we refer to as fluid dynamics grounding.
    \myitem We propose \model{}, the first differentiable model that attempts to understand the fluid dynamics by solving the inverse problem of synthesizing the visual scenes.
    \myitem We propose \render{}, which allows for joint optimization of dynamics propagation and rendering by geometrically correlating fluid particles with neural radiance fields.
\end{itemize}
\vspace{-4pt}
Code and video demonstrations are available at \url{https://syguan96.github.io/NeuroFluid}.

\section{Problem Formulation}
\label{sec:problem_setup}

We solve the problem of grounding the physical dynamics of particle-based fluid systems from a sequence of visual observations $\{I_t^{\boldsymbol{d}}\}_{t=1:T}$ collected from a sparse set of views $\{\boldsymbol{d}\}$.
The dynamics can be represented by a particle state transition model $\mathcal{T}_\theta$. 
Given the initial particles state $\boldsymbol{s}_{0}$, $\mathcal{T}_\theta$ estimates the transitions of particle states from $\boldsymbol{s}_t$ to $\boldsymbol{s}_{t+1}$ with probability $p(\boldsymbol{s}_{t+1}|\boldsymbol{s}_{t};\theta)$.
Unlike existing learning-based fluid simulators \cite{li2018learning,mrowca2018flexible,sanchez2020learning,li2020visual,schenck2018spnets,ummenhofer2019lagrangian,kim2019deep}, in our problem setup, the parameters $\theta$ \textbf{CANNOT} be optimized with ground-truth particle states.

Instead, we receive a new observation $I_{t+1}^{\boldsymbol{d}}$ that can be used to solve the inverse graphics problem of particle-based fluid rendering, that is, to analyze the underlying physical properties of visual scenes by learning a differentiable renderer $\mathcal{R}_\phi$ to synthesize them.
The forward modeling process of fluid dynamics grounding can be formulated as:
\begin{equation}
\label{eq:problem_forward}
\begin{split}
    \boldsymbol{s}_{t+1} &\leftarrow \mathcal{T}_\theta(\boldsymbol{s}_{t}), \\
    \widehat{I}^{\boldsymbol{d}}_{t+1} &\leftarrow \mathcal{R}_\phi(\boldsymbol{s}_{t+1}, \boldsymbol{d}),
\end{split}
\end{equation}
where $\widehat{I}^{\boldsymbol{d}}_{t+1}$ is the synthesized image at time $(t+1)$ with view direction $\boldsymbol{d}$.
The main objective of fluid dynamics grounding is to obtain optimal $\theta^\ast$ and $\phi^\ast$ that can maximize the log-likelihood function of the observed images:
\begin{equation}
\label{eq:problem_objective}
    \mathop{\arg\max}_{\theta,\phi} \sum_{t,\boldsymbol{d}} \log\big(p({I}^{\boldsymbol{d}}_{t+1}|\boldsymbol{s}_{t+1},\boldsymbol{d}; {\phi})  \ p(\boldsymbol{s}_{t+1}|\boldsymbol{s}_t;{\theta})\big).
\end{equation}
We evaluate $\mathcal{T}_\theta$ by (i) measuring the distance between the optimized particle positions $\{\widehat{\boldsymbol{P}}_t\}_{1:T}$ and the true positions $\{\boldsymbol{P}_t\}_{1:T}$.
Further, it can be partly assessed through forward modeling of (ii) novel view synthesis and (iii) rolling-out $\mathcal{T}_\theta$ iteratively to predict multiple steps into the future.

\section{\model{}}

In this section, we present the details of \model{}, which is an optimization-based approach that understands fluid physics by learning to synthesize sequences of visual observations. 
The overall framework in the forward modeling phase includes two steps: fluid simulation and dynamic scene rendering (see Figure~\ref{fig:intro}). They are in form of neural networks parametrized by $\theta$ and $\phi$ respectively.

In Section~\ref{subsec:dynamics}, we first describe the particle-based representations that connect fluid dynamics modeling with neural rendering and then describe the particle-based dynamics modeling approach in \model{}.
In Section~\ref{subsec:renderer}, we revisit the rendering principles of NeRF and present the particle-driven neural radiance fields that are specifically designed for fluid dynamic scenes. 
In Section~\ref{subsec:optim}, we discuss the optimization procedure of \model{}, which is the key to solving the inverse problem of rendering the physical scenes.
Notably, \model{} is the first fully differentiable model for fluid dynamics grounding.

\subsection{Particle-Based Dynamics Modeling}
\label{subsec:dynamics}

\paragraph{Why particle-based representations?}
Particle-based representations facilitate fluid dynamics grounding.
Fluid representation techniques have been fully explored in previous literature~\cite{ummenhofer2019lagrangian,li2019learning}, in which the particle-based approaches have the advantage of simulating complex fluid dynamics by modeling the interactions of a group of particles. 
Another advantage of particle-based representations is that they can easily back-propagate gradients, and they are therefore widely used in differentiable fluid simulators.

These properties make the particle-based representations particularly suitable for our optimization-based dynamics grounding approach: First, in the forward modeling process, they can intuitively reflect fluid dynamics and 3D geometries, and thus effectively drive the subsequent neural renderer. 
Second, during the optimization process, the particles estimated by the state transition model ($\mathcal{T}_\theta$) receive the gradients generated by the error of the rendering results,  allowing $\mathcal{T}_\theta$ to gradually approach the real fluid dynamics.

In practice, we explicitly specify the particle states $\boldsymbol{s}_t$ in Eq.~\eqref{eq:problem_forward} and Eq.~\eqref{eq:problem_objective} as the particle positions $\boldsymbol{P}_t$ and velocities $\boldsymbol{V}_t$ over the entire particle set.

\begin{figure*}[t]
\vskip 0.2in
\begin{center}
\centerline{\includegraphics[width=\linewidth]{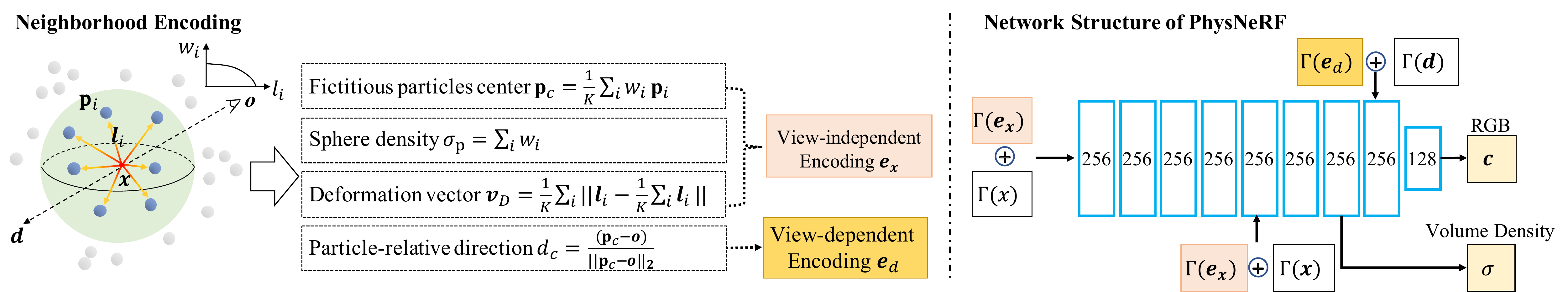}}
\vskip -0.1in
\caption{Overview of \render{}, a key component in \model{}.
\textbf{Left:} In \render{}, the physical properties of fluid particles are involved in the radiance fields through the proposed neighborhood encoding scheme.
\textbf{Right:} \render{} employs view-independent particle encoding $\boldsymbol{e}_{\boldsymbol{x}}$ and view-dependent particle encoding $\boldsymbol{e}_{d}$ to estimate the volume density $\sigma$ and the emitted color $\boldsymbol{c}$. 
}
\label{fig:renderer}
\end{center}
\vskip -0.2in
\end{figure*}

\myparagraph{Particle transition model.}

In forward modeling, the particle transition model $\mathcal{T}_{\theta}$ learns the movement and interactions of particles between consecutive time steps.
Starting with the initial particle states $(\boldsymbol{P}_{0}, \boldsymbol{V}_{0})$, which are known during training and testing phases, it roll-outs over time to estimate a sequence of particle states $\{(\boldsymbol{P}_{t}, \boldsymbol{V}_{t})\}_{1:T}$.
At each time step, $(\boldsymbol{P}_{t}, \boldsymbol{V}_{t})$ represent the underlying geometric information of the fluid; They are then fed into a particle-driven neural renderer to synthesize view-dependent images. 
A well-performed particle transition model can effectively predict multiple time steps into the future beyond the observed time scope during training time.

It is worth noting that, \model{} does not have special requirements for the transition model, in the sense that any particle-based fluid simulation network can plug and play in our framework as the particle transition model.
Without loss of generality, we here use a state-of-the-art differentiable fluid simulator named Deep Lagrangian Fluids (DLF)~\cite{ummenhofer2019lagrangian}.
It leverages an improved convolutional operator to capture the physical relations between the particles in small neighborhoods of 3D space, thus achieving a compromise between prediction accuracy and computational efficiency.

Due to the lack of ground-truth $\{(\boldsymbol{P}_{t}, \boldsymbol{V}_{t})\}_{1:T}$, the key challenge in dynamics grounding is to find an appropriate way to supervise $\mathcal{T}_{\theta}$ such that it can intuitively approximate the physical dynamics of the fluid. This appropriate way, in our case, is the particle-driven neural radiance fields.

\subsection{Particle-Driven Neural Radiance Fields}
\label{subsec:renderer}

\paragraph{Revisiting NeRF.} 
In the conventional approach of neural radiance fields (NeRF)~\cite{mildenhall2020nerf}, the color of each pixel on the rendered image is decided in the following steps. 
First, a ray is emitted from the camera location $\boldsymbol{o}$ along the view direction $\boldsymbol{d}$, which can be denoted by $\boldsymbol{r} = \boldsymbol{o}+n\boldsymbol{d}$. 
Then, $N$ points with 3D coordinates $\{\boldsymbol{x}_i\}_{1:N}$ are sampled along $\boldsymbol{r}$ between the near and far bounds of the physical scene.
Next, a multilayer perceptron (MLP) is trained to map $(\boldsymbol{x}_i, \boldsymbol{d})$ to the intrinsic volume density $\sigma(\boldsymbol{x}_i)$ and the view-dependent emitted color $\boldsymbol{c}(\boldsymbol{x}_i, \boldsymbol{d})$. 
Finally, the RGB value $C(\boldsymbol{r})$ is decided using a classic volume rendering function~\cite{kajiya1984ray}:
\begin{equation}
\label{eq:nerf_func}
\begin{split}
    C(\boldsymbol{r}) &= \sum_{i=1}^{N} T_{i} \Big(1-\exp\big(-\sigma(\boldsymbol{x}_i) \delta_i\big)\Big) \boldsymbol{c}(\boldsymbol{x}_i, \boldsymbol{d}), \\ 
    T_i &= \exp\left(-\sum_{j=1}^{i-1} \sigma(\boldsymbol{x}_j) \delta_j \right),
\end{split}
\end{equation}
where $\delta_j = \|\boldsymbol{x}_{j+1}-\boldsymbol{x}_j\|_2$ is the interval between two adjacent sampling points. Note that the positional encoding and hierarchical sampling strategy are omitted for brevity.
In our work, $C(\boldsymbol{r})$ is represented as $I_t^{\boldsymbol{d}}(u,v)$, where $(u,v)$ is the interaction between the ray and the image plane.

\myparagraph{PhysNeRF: Linking particles with radiance fields.} 
We consider how fluid particles are associated with the radiance field in neural rendering. The motivation is that \emph{a reasonable design of the renderer that conforms to physics can intuitively facilitate grounding fluid dynamics through joint optimization of particle transition and neural rendering.}

One hint is that the volume rendering result along a given ray is closely related to the geometric distribution of its neighboring physical particles. In extreme cases, the fewer particles around the ray, the closer the accumulated color $I(u,v)$ is to the background color.
We propose \render{} based on this assumption. 
Given a sampling point at $\boldsymbol{x}$, \render{} first conducts ball query~\cite{qi2017pointnetpp} to search its neighboring fluid particles  $\boldsymbol{P}^{\boldsymbol{x}} \in \boldsymbol{P}_t$:
\begin{align}
\label{eq:ball_query}
    \boldsymbol{P}^{\boldsymbol{x}} = \mathcal{S}(\|\boldsymbol{\mathrm{p}}_i - \boldsymbol{x}\|_2, r_s),
\end{align}
where $\boldsymbol{\mathrm{p}}_i$ refers to the $i$-th particle in $\boldsymbol{P}_t$, produced by $\mathcal{T}_\theta$ at time step $t$, and $r_s$ is the search radius. In line with the particle transition model from \cite{ummenhofer2019lagrangian}, we set $r_s=9 \cdot r_p$, where $r_p$ is radius of the fluid particle. 
Furthermore, we assume that both the view-independent volume density and view-dependent color of the sampling point should be dependent on the geometric properties $\boldsymbol{P}^{\boldsymbol{x}}$ in its 3D neighborhood.
Therefore, we parameterize $\boldsymbol{P}^{\boldsymbol{x}}$ to view-independent encoding $\boldsymbol{e}_{\boldsymbol{x}}$ and view-dependent encoding $\boldsymbol{e}_{\boldsymbol{d}}$ that can be written as follows:
\begin{align}
\label{eq:nn_encode}
    (\boldsymbol{e}_{\boldsymbol{x}}, \boldsymbol{e}_{\boldsymbol{d}}) = \mathcal{E}(\boldsymbol{P}^{\boldsymbol{x}}, \boldsymbol{x}),
\end{align}
where $\mathcal{E}(\cdot)$ refers to the parameterization operators, which will be described later.
Finally, we train an MLP network, denoted by $\mathcal{R}_\phi$, to map the compositional inputs $(\boldsymbol{e}_{\boldsymbol{x}}, \boldsymbol{e}_{\boldsymbol{d}})$ and $(\boldsymbol{x}_i, \boldsymbol{d})$ to volume density and emitted color. 
The rendering mechanism of \render{} can be formulated as follows:
\begin{align}
\label{eq:mlp_f}
    \big(\sigma(\boldsymbol{x}, \boldsymbol{e}_{\boldsymbol{x}}), \boldsymbol{c}(\boldsymbol{x}, \boldsymbol{e}_{\boldsymbol{x}}, \boldsymbol{d}, \boldsymbol{e}_{\boldsymbol{d}})\big) = \mathcal{R}_\phi(\boldsymbol{x}, \boldsymbol{d}, \boldsymbol{e}_{\boldsymbol{x}}, \boldsymbol{e}_{\boldsymbol{d}}).
\end{align}
Notably, the parameters of the MLP are shared at different time steps throughout the dynamic scene. With a fixed viewpoint, the differences in the rendering results are only determined by the geometric distribution of $\boldsymbol{P}_t$. It is the unique property of \render{} that makes it different from all existing NeRF-based neural renderers.

\myparagraph{Neighborhood encoding in \render{}.}
We here introduce how to encode the neighboring particles $\boldsymbol{P}^{\boldsymbol{x}}$ to $(\boldsymbol{e}_{\boldsymbol{x}}, \boldsymbol{e}_{\boldsymbol{d}})$.
As shown in Figure~\ref{fig:renderer}, we first define $\boldsymbol{\mathrm{p}}_c$ as the weighted average position over $\boldsymbol{P}^{\boldsymbol{x}}$:
\begin{align}
    \boldsymbol{\mathrm{p}}_c = \frac{1}{K} \sum_{i} w_{i} \boldsymbol{\mathrm{p}}_{i}, \quad \boldsymbol{\mathrm{p}}_{i} \in \boldsymbol{P}^{\boldsymbol{x}},
\end{align}
where $K$ is the number of neighboring particles. We use $w_{i}$ to indicate the contribution weight of each particle to form the fictitious center particle, which is calculated by
\begin{align}
    w_{i} = \max\Big(1-\big(\frac{\|\boldsymbol{l}_{i}\|_2}{r_s}\big)^3, 0\Big),
\end{align}
where $\boldsymbol{l}_{i} = \boldsymbol{\mathrm{p}}_i-\boldsymbol{x}$ is the local vector from the sampling point $\boldsymbol{x}$ to 
particle $\boldsymbol{\mathrm{p}}_i$. Intuitively, $w_{i}$ describes the fact that the closer $\boldsymbol{\mathrm{p}}_i$ is to the sampling point $\boldsymbol{x}$, the greater its contribution to the rendering result at $\boldsymbol{x}$. 
Having the fictitious center particle $\boldsymbol{\mathrm{p}}_c$, we calculate the normalized view direction from the camera location $\boldsymbol{o}$ to $\boldsymbol{\mathrm{p}}_c$, which is an importance reference direction for network to infer ray refraction and reflection, as $\boldsymbol{d}_{c} = (\boldsymbol{\mathrm{p}}_c - \boldsymbol{o})/\|\boldsymbol{\mathrm{p}}_c - \boldsymbol{o}\|_2$.

To infer the volume density $\sigma$ at $\boldsymbol{x}$, a straightforward idea is to use the number of particles in $\boldsymbol{P}^{\boldsymbol{x}}$ as a closely related physical condition.
Nevertheless, to avoid optimizing a discrete variable, we define a soft version of the particle density around $\boldsymbol{x}$ by calculating $ \sigma_\mathrm{p} = \sum_{i} w_{i}$.
It meets the fact that the closer of $\boldsymbol{\mathrm{p}}_{i}$ is to sampling point $\boldsymbol{x}$, the less likely the ray is to pass through $\boldsymbol{x}$.

Although $\boldsymbol{P}^{\boldsymbol{x}}$ is searched by an isotropic sphere, the actual particle distribution within it is an-isotropic. Partly inspired by the work from \citet{biedert2018direct}, we calculate a radial vector to represent the deformation in the particle set:
\begin{align}
    \boldsymbol{v}_\mathrm{D} = \frac{1}{K} \sum_{i} \big\|\boldsymbol{l}_{i} - \frac{1}{K}\sum_{i} \boldsymbol{l}_{i} \big\|_2.
\end{align}
Finally, we take into account all of the above physical quantities and derive the view-independent encoding and the view-dependent encoding as
\begin{align}
    \boldsymbol{e}_{\boldsymbol{x}} = \big(\Gamma(\boldsymbol{\mathrm{p}}_c), \Gamma(\sigma_\mathrm{p}), \Gamma(\boldsymbol{v}_\mathrm{D})\big), \quad \boldsymbol{e}_{\boldsymbol{d}} = \Gamma(\boldsymbol{d}_c),
\end{align}
where $\Gamma(\cdot)$ refers to the positional encoding functions in NeRF~\cite{mildenhall2020nerf}.

\subsection{Optimizing \model{}}
\label{subsec:optim}

Similar to NeRF, at a specific time step, we use the mean squared error to constrain the rendering result:
\begin{align}
    \mathcal{L}(\widehat{I}_t^{\boldsymbol{d}}, {I}_t^{\boldsymbol{d}}) = \sum_{u,v} \big\|\widehat{I}_t^{\boldsymbol{d}}(u,v) - I_t^{\boldsymbol{d}}(u,v) \big\|_2^2,
\end{align}
where $I_t^{\boldsymbol{d}}(u,v)$ is typically represented as $C(\boldsymbol{r})$ in NeRF.

Since we generally do not have any prior knowledge of the physical properties of the fluid, we can hardly impose further constraints on particle states. 
Under these circumstances, jointly training $\mathcal{T}_{\theta}$ and $\mathcal{R}_{\phi}$ from a cold start makes the optimization prone to collapse to some undesirable local minima.
Therefore, we first warm-up \render{} on the initial fluid scene with a sparse set of $N_{\boldsymbol{d}}$ views conditioned on $(\boldsymbol{P}_0, \boldsymbol{V}_0)$ to obtain an effective initialization of the renderer. The objective function in the warm-up phase is 
\begin{align}
\label{eq:obj1}
    \mathop{\text{minimize}}\limits_{\phi} \ \sum_{{\boldsymbol{d}}} \mathcal{L}({\widehat{I}_0^{\boldsymbol{d}}}, {I_0^{\boldsymbol{d}}}; \boldsymbol{P}_0, \boldsymbol{V}_0, \phi).
\end{align}

After warming-up \render{}, it roughly learns the correlation between view-dependent observations and particle geometries, we then jointly train $\mathcal{T}_{\theta}$ and $\mathcal{R}_{\phi}$ on $\{I_t^{\boldsymbol{d}}\}_{t=1:T}$, collected from a static camera with a fixed direction of $\boldsymbol{d}$. The objective function in this end-to-end training phase is
\begin{align}
\label{eq:obj2}
    \mathop{\text{minimize}}\limits_{\theta,\phi} \ \sum_{t=1}^{T} \mathcal{L}({\widehat{I}_t^{\boldsymbol{d}}}, {I_t^{\boldsymbol{d}}};\boldsymbol{P}_0, \boldsymbol{V}_0, \theta,\phi).
\end{align}

\begin{table*}[t]
\caption{Typical geometric and physical properties of fluids on the evaluation benchmarks, which are closely related to the simulation and rendering of dynamic scenes. On ``WaterBunny'', we evaluate the generalization ability of \render{} to novel particle distributions. 
}
\setlength\tabcolsep{10pt}
\label{tab:benckmark}
\vskip 0.1in
\begin{center}
\begin{small}
\begin{sc}
\begin{tabular}{lcccc}
\toprule
Benchmark & Initial shape & Material & Viscosity & Density (kg/m$^3$) \\
\midrule
{HoneyCone} & Cone & Principled BSDF & 0.8 & 1420 \\
{WaterCube} & Cube & Glass BSDF & 0.08 & 1000 \\
{WaterSphere} & Sphere & Glass BSDF & 0.08 & 1000 \\
\midrule
WaterBunny & StanfordBunny & Glass BSDF & 0.08 & 1000 \\
\bottomrule
\end{tabular}
\end{sc}
\end{small}
\end{center}
\vskip -0.1in
\end{table*}

\myparagraph{Implementation and training details.} 
The model architecture of \render{} is the same as NeRF, except that it takes as input $(\boldsymbol{e}_{\boldsymbol{x}}, \boldsymbol{e}_{\boldsymbol{d}})$.
We adopt the hierarchical sampling strategy to sample $64$ points coarsely and then sample $128$ points finely along each ray.
The positional encoding parameters for $\boldsymbol{e}_{\boldsymbol{x}}$ and $\boldsymbol{e}_{\boldsymbol{d}}$ are the same as those for $\boldsymbol{x}$ and $\boldsymbol{d}$.
\model{} are trained with the Adam optimizer \cite{kingma2015adam}.
In the warm-up phase of $\mathcal{R}_{\phi}$, it is trained for $100$k steps with an initial learning rate of $5e^-4$, which is then exponentially decayed with $\gamma=0.1$.
In the joint training phase, the entire model is trained for $500$k steps. The initial learning rate of $\mathcal{T}_{\theta}$ and $\mathcal{R}_{\phi}$ are set to $1e^{-6}$ and $5e^{-4}$ respectively.
The learning rate of $\mathcal{T}_{\theta}$ is decayed by $0.5$ after $10$k, $30$k, $50$k, $100$k, and $300$k steps. That of $\mathcal{R}_{\phi}$ is decayed by $0.5$ after $10$k, $75$k, and $150$k steps. 
To ease the convergence, we initialize $\mathcal{T}_{\theta}$ with the released DLF model that is pre-trained on other benchmarks. The resolution of the observed image is $400 \times 400$.

\section{Experiments}

\subsection{Experimental Setup}

The effect of fluid dynamics grounding is evaluated from three aspects: distances between the grounded and true particle positions (Section~\ref{sec:trans}), distances between the predicted particles and the ground truth in a future time horizon (Section~\ref{sec:rollout}), and the quality of novel view synthesis (Section~\ref{sec:novelview}).

\myparagraph{Benchmarks.}
We train \model{} on visual observations of fluids generated by DFSPH~\cite{bender2015divergence} and Blender~\cite{blender}.
DFSPH is a particle-based physics engine that can simulate complex fluid dynamics. 
Blender is used to render the simulated particles with different materials to produce high-fidelity and realistic images. 
We generate four benchmarks named ``HoneyCone'', ``WaterCube'', ``WaterSphere'', and ``WaterBunny''.
We simulate the fluid dynamics falling inside a cube box for $60$ time steps.
As shown in Table~\ref{tab:benckmark}, different benchmarks vary in the initial shape, material, viscosity, and density of the fluid.
For HoneyCone, WaterCube, and WaterSphere, we use the visual observations during the first $50$ time steps as the training data.
In particular, we use WaterBunny to evaluate the generalization ability of \render{}, which is pre-trained on the first three benchmarks and evaluated on novel particle distributions of StanfordBunny for image synthesis.

\myparagraph{Evaluation metrics.}
Following existing fluid simulation approaches~\cite{ummenhofer2019lagrangian}, we use the Euclidean distance to measure the performance of particle grounding and future prediction, which is calculated by $d = \frac{1}{N} \sum_{i=1}^{N} \min_{{\boldsymbol{\mathrm{p}}}_i \in {\boldsymbol{P}}_t} \|\boldsymbol{\mathrm{p}}_i - \widetilde{\boldsymbol{\mathrm{p}}}_i\|_2$, where ${\boldsymbol{P}}_t$ is the set of estimated particles at a specific time step, and $\widetilde{\boldsymbol{\mathrm{p}}}_i$ is the ground-truth particle position for ${\boldsymbol{\mathrm{p}}}_i$. 
To evaluate the quality of novel view synthesis, we follow the original NeRF \cite{mildenhall2020nerf} to use PSNR, SSIM~\cite{wang2004image}, and LPIPS~\cite{zhang2018unreasonable} as the metrics. Higher PSNR and SSIM values and a lower LPIPS indicate better rendering of fluid scenes.

\begin{table*}[t]
\caption{Quantitative results on the errors of fluid dynamics grounding ($t<50$) and prediction ($50\le t<60$), which are calculated between the grounded/predicted particle positions and the ground truth provided by the fluid simulator (lower is better).
For \textbf{DLF$^\dagger$}, the transition model is finetuned on the testing benchmarks in a fully supervised way, that is, using \textbf{true} particle positions at $t<50$.}
\label{tab:grounding_prediction}
\vskip 0.1in
\begin{center}
\begin{small}
\begin{sc}
\begin{tabular}{lcccccccccccc}
\toprule
\multirow{3}{*}{Method}
& \multicolumn{4}{c}{WaterCube}
& \multicolumn{4}{c}{WaterSphere}
& \multicolumn{4}{c}{HoneyCone}
\\
& \multicolumn{2}{c}{Grounding} 
& \multicolumn{2}{c}{Prediction} 
& \multicolumn{2}{c}{Grounding} 
& \multicolumn{2}{c}{Prediction} 
& \multicolumn{2}{c}{Grounding} 
& \multicolumn{2}{c}{Prediction} 
\\
& $d_{t<50}^{\text{avg}}$ &$d_{t=49}$ & $d^{\text{avg}}_{t\ge50}$ &$d_{t=59}$
& $d_{t<50}^{\text{avg}}$ &$d_{t=49}$ & $d^{\text{avg}}_{t\ge50}$ &$d_{t=59}$
& $d_{t<50}^{\text{avg}}$ &$d_{t=49}$ & $d^{\text{avg}}_{t\ge50}$ &$d_{t=59}$ \\
\cmidrule(lr){1-1}          \cmidrule(lr){2-5}  \cmidrule(lr){6-9}  \cmidrule(lr){10-13}
DLF                   
&32.3 &48.3 &47.4 &46.2         
&32.2 &47.6 &48.1 &45.9        
&61.5 &83.5 & 69.7 &\textbf{57.8}\\
\model{}                
&\textbf{28.8} &\textbf{34.9} &\textbf{35.5} &\textbf{36.7}         
&\textbf{31.1} &\textbf{31.5} &\textbf{30.7} &\textbf{30.4}          
&\textbf{30.9} &\textbf{47.5} &\textbf{54.2} & {58.2}
\\
\cmidrule(lr){1-13}     
{DLF$^\dagger$}         
&{28.1} &{28.1} &{30.9} &{34.4} 
&{30.0} &{28.5} &{30.0} &{31.8}  
&{34.3} &{66.1} &72.6 &77.6\\
\bottomrule
\end{tabular}
\end{sc}
\end{small}
\end{center}
\end{table*}

\begin{figure*}[t]
\vskip 0.1in
\begin{center}
\centerline{
\includegraphics[width=0.95\linewidth]{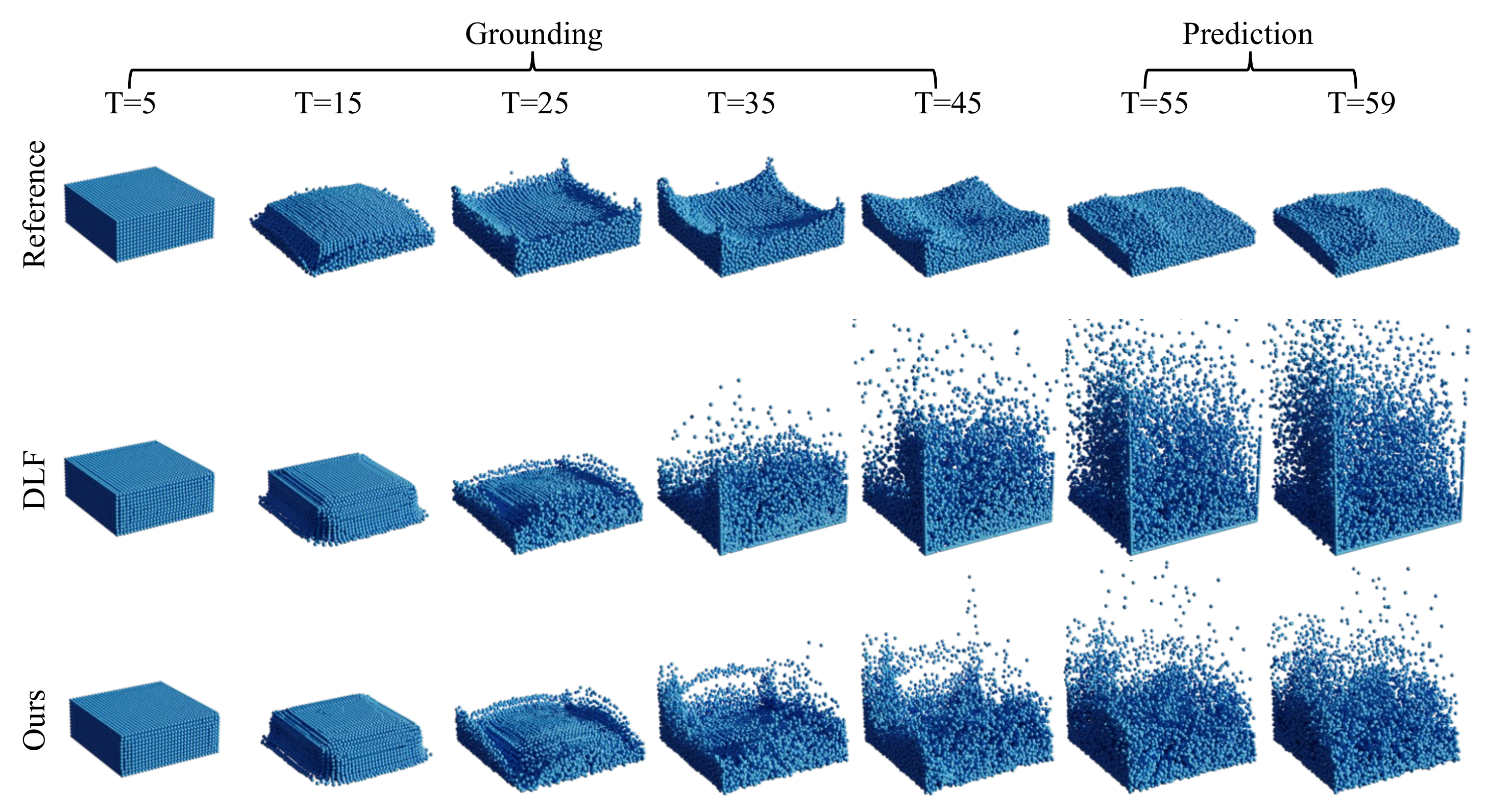}
}
\vspace{-5pt}
\caption{Qualitative results of fluid dynamics grounding the future dynamics prediction. From the top to bottom, we visualize ground-truth particles, particles predicted by DLF, and particles grounded/predicted by \model{}.}
\label{fig:particle_vis}
\end{center}
\vskip -0.2in
\end{figure*}

\begin{figure*}[t]
\vskip 0.2in
\begin{center}
\centerline{
\includegraphics[width=0.9\linewidth]{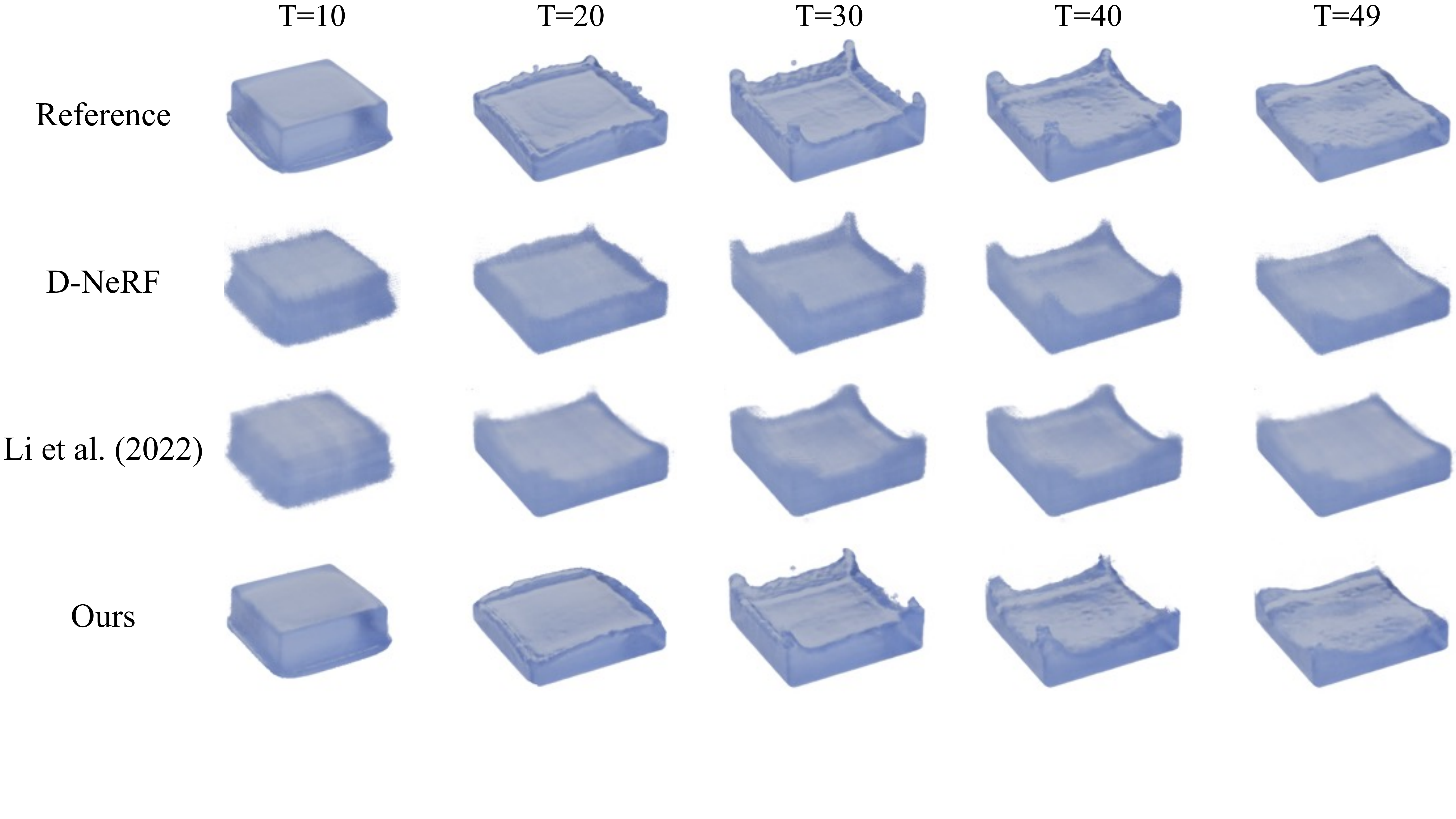}
}
\vspace{-45pt}
\caption{
Novel view synthesis on WaterCube. The first row shows the reference image sequence provided by the Blender renderer. The synthesized images of \model{} better preserve the fluid dynamics and fine details of the fluid surface. This indicates that \render{} can effectively incorporate the geometry of fluid particles.
}
\label{fig:novelview}
\end{center}
\vskip -0.2in
\end{figure*}

\begin{table*}[t]
\caption{Quantitative results of novel view synthesis averaged over the observation period ($t\in [0, 49]$).}
\label{exp:ssim}
\vskip 0.1in
\begin{center}
\begin{small}
\begin{sc}
\begin{tabular}{lcccccccccccc}
\toprule
\multirow{2}{*}{Method}
& \multicolumn{3}{c}{WaterCube} 
&\multicolumn{3}{c}{WaterSphere} 
&\multicolumn{3}{c}{HoneyCone}\\
& PSNR$\uparrow$  & SSIM$\uparrow$ & LPIPS$\downarrow$
& PSNR$\uparrow$  & SSIM$\uparrow$ & LPIPS$\downarrow$
& PSNR$\uparrow$  & SSIM$\uparrow$ & LPIPS$\downarrow$\\
\cmidrule(r){1-1}  \cmidrule(lr){2-4}  \cmidrule(r){5-7}  \cmidrule(r){8-10}
NeRF            &21.36&0.90&0.31    &19.61&0.81&0.39    &23.33&0.92&0.22\\
NeRF-T          &24.04&0.90&0.26    &21.58&0.89&0.35    &23.95&0.92&0.22\\
D-NeRF          &30.08&0.93&0.16    &28.80&0.92&0.21    &31.02&0.95&0.13\\
\citet{li20223d} &26.13&0.93&0.15    &21.81&0.89&0.35    &15.88&0.87&0.36\\
\model{}         &\textbf{30.76}&\textbf{0.95}&\textbf{0.09}   &\textbf{33.24}&\textbf{0.96}&\textbf{0.10}  &\textbf{37.82}&\textbf{0.99}&\textbf{0.05}\\
\bottomrule
\end{tabular}
\end{sc}
\end{small}
\end{center}
\vskip -0.1in
\end{table*}

\begin{table*}[t]
\caption{Quality of image rendering for the next $10$ time steps ($t\in [50, 59]$).}
\label{tab:dynamics_quan_exp_vs}
\vskip 0.1in
\begin{center}
\begin{small}
\begin{sc}
\begin{tabular}{lcccccccccccc}
\toprule
\multirow{2}{*}{Method}
& \multicolumn{3}{c}{WaterCube} 
&\multicolumn{3}{c}{WaterSphere} 
&\multicolumn{3}{c}{HoneyCone}\\
& PSNR$\uparrow$  & SSIM$\uparrow$ & LPIPS$\downarrow$
& PSNR$\uparrow$  & SSIM$\uparrow$ & LPIPS$\downarrow$
& PSNR$\uparrow$  & SSIM$\uparrow$ & LPIPS$\downarrow$\\
\cmidrule(r){1-1}  \cmidrule(lr){2-4}  \cmidrule(r){5-7}  \cmidrule(r){8-10}
NeRF                &21.46&0.90&0.30 &18.75&0.80&0.39 &20.65&0.91&0.23\\
NeRF-T              &24.06&0.89&0.27 &20.67&0.87&0.40 &21.37&0.90&0.23\\
D-NeRF              &27.28&0.92&0.17 &26.95&0.90&0.22 &23.25&0.92&0.16\\
\citet{li20223d}     &26.96&0.93&0.15 &21.22&0.89&0.35 &16.09&0.87&0.35 \\
\model{}            &\textbf{28.50}&\textbf{0.93}&\textbf{0.13} &\textbf{28.69}&\textbf{0.93}&\textbf{0.14} &\textbf{29.69}&\textbf{0.96}&\textbf{0.09} \\
\bottomrule
\end{tabular}
\end{sc}
\end{small}
\end{center}
\vskip -0.1in
\end{table*}

\begin{table*}[t]
\caption{Experiments on WaterCube with unknown initial particle states and ablation studies of neighborhood encoding (Rows 3-6). 
}
\label{tab:nnscheme_ablates}
\vskip 0.1in
\setlength\tabcolsep{8pt}
\begin{center}
\begin{small}
\begin{sc}
\begin{tabular}{lccccccc}
\toprule
\multirow{2}{*}{Model}
& \multicolumn{2}{c}{Grounding} & \multicolumn{2}{c}{Prediction} & \multicolumn{3}{c}{Novel view synthesis} 
\\
& $d_{t<50}^{\text{avg}}$ &$d_{t=49}$ & $d^{\text{avg}}_{t\ge50}$ &$d_{t=59}$
& PSNR$\uparrow$  & SSIM$\uparrow$ & LPIPS$\downarrow$ \\
\cmidrule(r){1-1}  \cmidrule(lr){2-3}  \cmidrule(r){4-5}  \cmidrule(r){6-8}
Full Model                                 &\textbf{28.8} &\underline{34.9} &\underline{35.5} &\underline{36.7} &\textbf{30.76} &0.95 &\textbf{0.09}   \\
Unknown initial particle positions       &35.6 &\textbf{27.2} &\textbf{26.6} &\textbf{26.3} &29.21 &0.94 &0.12 \\
\textit{w/o} Fictitious particles center ($\boldsymbol{\mathrm{p}}_c$)  &37.2 &40.7 &41.3 &42.9 &28.41 &0.94 &0.12   \\
\textit{w/o} Sphere density ($\sigma_\mathrm{p}$)             & \underline{31.2} &37.9 &39.3 &39.4 &\underline{29.65} &0.95 &\underline{0.10}    \\
\textit{w/o} Deformation vector ($\boldsymbol{v}_\mathrm{D}$)           &33.0 &38.1 &40.5 &42.1 &28.91 &0.95 &0.11    \\
\textit{w/o} Particle-relative direction ($\boldsymbol{d}_{c}$)  &32.2 &39.8 &43.9 &47.0 &29.56 &0.95 &\underline{0.10}    \\
\bottomrule 
\end{tabular}
\end{sc}
\end{small}
\end{center}
\vskip -0.1in
\end{table*}

\myparagraph{Compared methods for particle grounding and future prediction.}
Since \model{} is a pilot study for inferring fluid dynamics from visual observations alone, it is difficult to make a completely fair comparison with existing approaches. 
Therefore, we make a compromise by comparing it with two baseline models for fluid simulation.
Like the particle transition model in \model{}, they follow the same DLF architecture~\cite{ummenhofer2019lagrangian}. Given the initial particle positions and velocities, DLF uses a continuous convolution network that is performed in 3D space to simulate the particle transitions.
However, theses two baseline models are pre-trained in different scenarios:
\begin{itemize}[leftmargin=*]
    \vspace{-5pt}
    \item \textbf{DLF:} This model has the same network parameters as the initialization of the particle transition model in \model{}. It is well-trained on the benchmarks used in \cite{ummenhofer2019lagrangian}, supervised with true particle positions. Although DLF has shown the generalization ability to a wide variety of fluid shapes, materials, and environments, in \model{}, it can be further improved by learning from visual observations of the fluid surface.
    \vspace{-4pt}
    \item \textbf{DLF$^\dagger$:} To provide DLF with stronger supervisions, we finetune the above DLF model with true particle states on the evaluation benchmarks in Table \ref{tab:benckmark}. The finetuning stage lasts for $50$ epochs with a learning rate of $1e^{-6}$.
\end{itemize}
\vspace{-5pt}
Note that although VGPL~\cite{li2020visual} can also infer particles positions from visual observation, it requires a well-trained and \textit{fixed} particle transition model. In other words, unlike \model{}, the goal of VGPL is not to learn the underlying dynamics from the observed fluid scenes.

\myparagraph{Compared methods for novel view synthesis.}
To demonstrate that \model{} obtains excellent rendering effects for fluid scenes by explicitly considering the particle distributions, we compare it with the NeRF-based approaches, including D-NeRF~\cite{pumarola2021d}, NeRF-T, and the 3D-aware fluid renderer from \citet{li20223d}:
\begin{itemize}[leftmargin=*]
    \vspace{-5pt}
    \item \textbf{NeRF:} It represents a static scene as 5D radiance fields of locations and view directions.
    \vspace{-4pt}
    \item \textbf{D-NeRF:} It extends NeRF to dynamic scenes with a deformation network that estimates 3D transitions $\Psi: (\boldsymbol{x}, t) \rightarrow \Delta \boldsymbol{x}$ in a canonical space. Same as their setting, we randomly select one view from four views at every time step. The candidate views are the same as those used in the warm-up training stage.
    \vspace{-4pt}
    \item \textbf{NeRF-T:} Referring to D-NeRF, NeRF can be extended on dynamic scenes represented with an additional time input, deriving a 6D radiance field of $(\boldsymbol{x}, \boldsymbol{d}, t)$.
    \vspace{-4pt}
    \item \textbf{3D-aware fluid renderer:} \citet{li20223d} used an image encoder to learn latent fluid states, and fed the latent states to a NeRF to render the fluid surface. Unlike \model{}, it is not based on the particles. Besides, it learns the dynamics model and the NeRF-based renderer separately. We take $4$ views for the training data, which are also involved in the warm-up training stage of \render{}.
    \vspace{-5pt}
\end{itemize}

\subsection{Performance on Fluid Dynamics Grounding} \label{sec:trans}

We compare \model{} for particle dynamics grounding with DLF and DLF$^\dagger$, which have the same architecture as the transition model in \model{} but are trained under the supervision of ground-truth particles.
We evaluate: (i) the averaged Euclidean distance of the estimated particles to ground truth during the observation period (\textit{i.e.},  $d^{\text{AVG}}_{t<50}$), (ii) the end-point particle distance at $t=49$.
All models are fed with true particle positions and velocities at $t=0$.

From Table~\ref{tab:grounding_prediction}, we can see that \model{} consistently performs better than DLF on all benchmarks, indicating that it effectively exploits visual observations to improve the accuracy of fluid dynamics grounding.
Besides, since DLF$^\dagger$ is trained with ground-truth particle positions right on the evaluation benchmarks, it is reasonable to obtain accurate estimations of particle positions. 
Even so, \model{} achieves comparable results with DLF$^\dagger$, especially in $d^{\text{AVG}}_{t<50}$.
In particular, \model{} significantly outperforms DLF$^\dagger$ on HoneyCone, whose viscosity is much lower than the training data of the pre-trained DLF model. This verifies that \model{} with the visual supervision can better adapt to the changes of fluid physical properties. 
Figure~\ref{fig:particle_vis} provides the corresponding visualization of the grounded particles, from which we can observe that \model{} learns more reasonable fluid dynamics than DLF.

\subsection{Performance on Future Prediction}\label{sec:rollout}

To study whether the transition model has learned intrinsic physical transition principles, we conduct forward prediction for the next $10$ time steps. 
Same as above, in Table~\ref{tab:grounding_prediction}, we compute the averaged Euclidean distance to future true particle positions ($d^{\text{AVG}}_{t\ge 50}$) and the end-point distance ($d_{t=59}$). 
Overall, the predicted fluid dynamics of \model{} is more accurate than DLF.
On WaterCube and WaterSphere, DLF$^\dagger$ performs slightly better than \model{}. 
On HoneyCone, however, the predicted particles of DLF$^\dagger$ diverge the most from the ground truth. A possible reason is that DLF$^\dagger$ is prone to overfit the training particle dynamics in the observation period, being supervised by true particle states.

\subsection{Performance on Novel View Synthesis} \label{sec:novelview}

In Figure~\ref{fig:novelview}, we study \render{} for novel view synthesis results at the observed time steps. 
Being fed with time-independent input, NeRF produces images that are static and blurry. 
Other compared methods can synthesize images with time-varying texture by simply encoding time index as input (NeRF-T), learning spatial deformation field (D-NeRF), or learning temporal latent representations \cite{li20223d}.
However, the artifacts of blurry textures and 3D geometry still exist.
In contrast, \model{} synthesizes photo-realistic images in novel views, as the neighborhood encoding scheme introduces particle-based physical guidance.
The rendered images show fine details of the fluid surface (\textit{e.g.}, droplets), and are highly consistent with the fluid dynamics of the reference sequence.

Table~\ref{exp:ssim} gives the quantitative results of novel view synthesis, where \model{} outperforms the compared neural renderers significantly in all metrics, especially in LPIPS, which is more in line with human perception \cite{zhang2018unreasonable}.

In Table~\ref{tab:dynamics_quan_exp_vs}, \model{} also performs best in image synthesis over a short snippet of future time horizon. 
This, in turn, validates the effectiveness of \model{} in learning rational fluid dynamics, thus benefiting the prediction of future particle states.

\subsection{Model Analysis}

\paragraph{Unknown initial particle positions.} 
In previous experiments, we use ground-truth particle states at the initial time step ($t=0$). 
Here we investigate what if \model{} is fed with estimated initial positions of the fluid particles.
Specifically, we estimate the initial positions as follows. First, we train an original NeRF model at the initial time step using the sparse views in the warm-up stage.
Second, we extract meshes from the results using the Marching Cube method.
Third, we sample particles in the volume surrounded by the meshes as the initial particles of \model{}. 
From Table~\ref{tab:nnscheme_ablates}, we can observe that even without the ground-truth initial particle positions, \model{} can produce decent results in dynamics grounding, prediction, and novel view synthesis.

\myparagraph{Ablation studies on neighborhood encoding.} 
To verify the effectiveness of the neighborhood encoding scheme, we experiment with different variants of \model{} by removing each component in \render{}. 
In Table~\ref{tab:nnscheme_ablates} (Rows 3-6), we report the performance of fluid dynamics grounding, future prediction, and novel view synthesis on WaterCube.
As shown, the full neighborhood encoding scheme outperforms all these baseline models, especially in dynamics grounding and prediction.
These results illustrate that the proposed representations of fictitious particle center ($\boldsymbol{\mathrm{p}}_c$), sphere density ($\sigma_\mathrm{p}$), deformation ($\boldsymbol{v}_\mathrm{D}$), and particle-relative direction ($\boldsymbol{d}_{c}$) are all essential for \model{}.

\begin{figure}[t]
\begin{center}
\centerline{
\includegraphics[width=0.85\columnwidth]{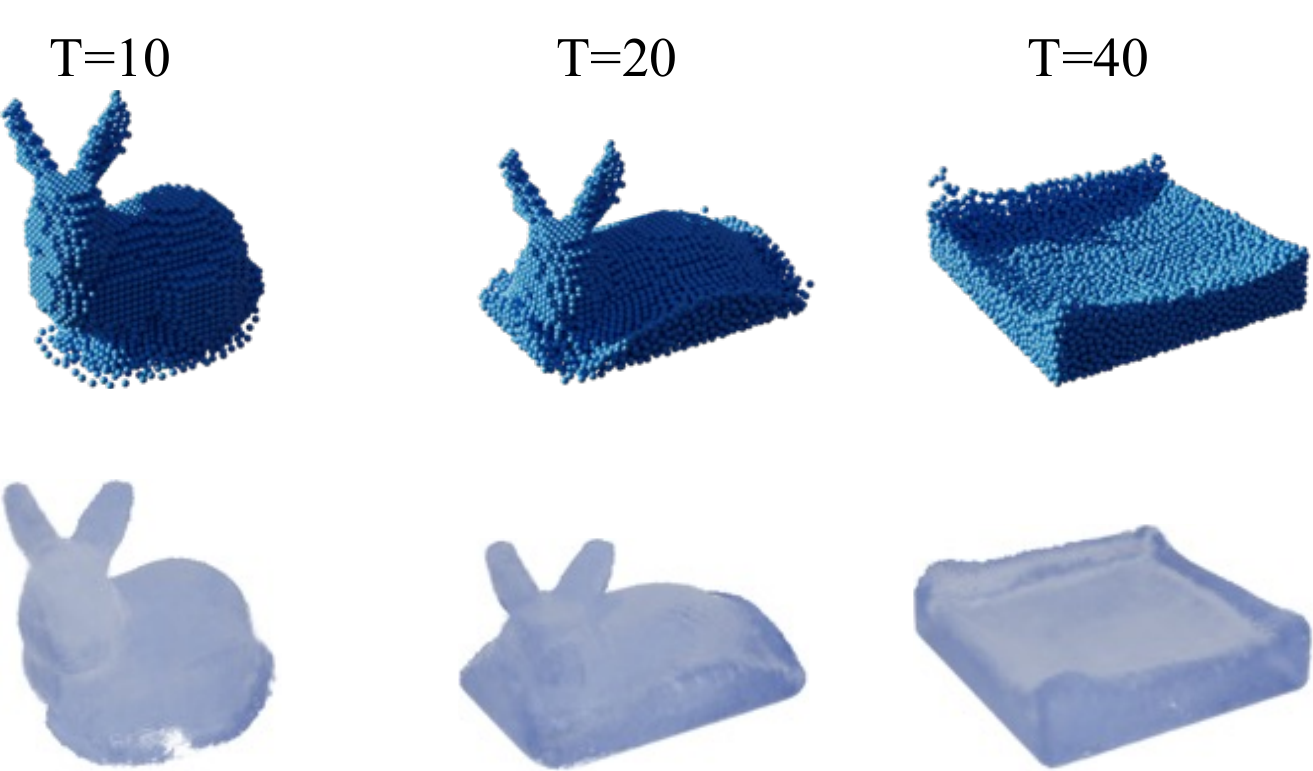}
}
\vspace{-5pt}
\caption{Results of \render{} conditioned on true particle positions of novel scenes (WaterBunny), showing effective rendering of details in lighting and fluid geometry.}
\label{fig:novel_particle_dynamics}
\end{center}
\vskip -0.2in
\end{figure}

\begin{table}[t]
\caption{Influence of image resolution for the joint training phase. All models are trained for $100$k steps on WaterCube. The proposed \model{} receives $400\times 400$px images during training.
}
\label{tab:resolution}
\vskip 0.1in
\setlength\tabcolsep{9pt}
\begin{center}
\begin{small}
\begin{sc}
\begin{tabular}{lccccccc}
\toprule
\multirow{2}{*}{Image Res.}
& \multicolumn{2}{c}{Grounding} & \multicolumn{2}{c}{Prediction}
\\
& $d_{t<50}^{\text{avg}}$ &$d_{t=49}$ & $d^{\text{avg}}_{t\ge50}$ &$d_{t=59}$
\\
\cmidrule(r){1-1}  \cmidrule(lr){2-3}  \cmidrule(r){4-5}
$200\times 200$ px     &29.2 &35.9 &37.2 &37.0  \\ 
$400\times 400$ px     &29.0 &35.1 &36.2 &36.0  \\ 
$800\times 800$ px     &29.0 &34.5 &36.0 &36.0  \\ 
\bottomrule 
\end{tabular}
\end{sc}
\end{small}
\end{center}
\vskip -0.1in
\end{table}

\myparagraph{Rendering novel fluid scenes.}
One contribution of \render{} is to use neighborhood encoding to correlate the particles with the neural radiance field.  
To verify that \render{} can effectively respond to different particle-based geometries, we use a pre-trained \render{} to render a novel fluid scene with a new initial shape of water, \textit{i.e.}, Stanford Bunny.
It is worth noting that the \render{} is trained on other benchmarks of water scenes with different initial shapes of cubes, spheres, and cones.
As shown in Figure~\ref{fig:novel_particle_dynamics}, when fed with the ground-truth particles of WaterBunny, \render{} renders the water surface with high fidelity, showing the generalization ability to unseen fluid shapes.
Because of this, \render{} can effectively drive the entire model to reason about fluid dynamics in end-to-end optimization.

\myparagraph{The resolution of observed images.} 
The proposed
\model{} receives $400 \times 400$px images in the training stage. In Table~\ref{tab:resolution}, we show the influence of changing the size of visual observations. 
It is important to note that the higher resolution can slightly improve the results of dynamics grounding and future prediction, while we can still obtain reasonable results using lower resolution images.

\section{Related Work}

\paragraph{Intuitive physics.}
Recent work in intuitive physics and inverse graphics has attempted to build neural models that can reason about 3D structures, stability, collisions, forces, and velocities from images or videos~\cite{battaglia2013simulation,wu2017learning,marrnet,RenderNet2018,handrwr2020,chen2021dib}.
The most relevant work to fluid dynamics grounding is \cite{li20223d}.
The difference, however, is that the model learns fluid dynamics on latent states encoded from visual observations, whereas our model infers the fluid dynamics in explicit particle space.

\myparagraph{Particle-based fluid simulation.}
Fluid simulation is a long-standing research field in computer graphics and related scientific areas~\cite{chorin1968numerical,stam1999stable,chentanez2013mass,sulsky1995application,zhu2005animating,stomakhin2013material,jiang2015affine,fang2020iq}. 
Particle-based methods represent fluid dynamics through the interactions of a group of particles~\cite{monaghan1992smoothed,muller2003particle,price2012smoothed}. They can naturally simulate complex collisions and be easily integrated into interactive physical environments~\cite{muller2003particle,amada2004particle} and differentiable simulators~\cite{hu2019difftaichi,holl2020learning}.
Learning fluid simulators from data greatly improves the efficiency of predicting complex fluid dynamics~\cite{muller1999application,poloni2000hybridization,ladicky2015data}.
Recent advances in deep learning typically use graph networks~\cite{li2018learning,mrowca2018flexible,sanchez2020learning,li2020visual} or modified convolution operators~\cite{schenck2018spnets,ummenhofer2019lagrangian,kim2019deep} to simulate particle state transitions and interactions in local neighborhoods.
However, these models need to be trained with consecutive true particle states, rather than reasoning about fluid dynamics directly from observed images, as \model{} does.

\myparagraph{Differentiable rendering.} 
Our fluid renderer is designed on top of Neural Radiance Fields (NeRF)~\cite{mildenhall2020nerf}, a technique that represents 3D scenes as learnable continuous functions. 
Different from other neural renderers, including mesh-based rasterizators~\cite{loper2014opendr,kato2018neural,liu2019soft} and ray tracers~\cite{li2018differentiable}, we find that NeRF is more compatible with particle-based fluid representations.
Existing approaches extend NeRF to deformed scenes with strong inductive biases of canonical displacement~\cite{pumarola2021d,park2021nerfies},
scene flow warping~\cite{li2021neural}, or even 3D human meshes and poses~\cite{peng2021animatable,liu2021neural,su2021nerf}, which do not perfectly match the geometric properties of particle systems. 
In contrast, \render{} is a particle-driven neural renderer and thus can facilitate fluid dynamics grounding through end-to-end optimization.

\section{Conclusions and Future Directions}

We studied a new research problem named fluid dynamics grounding, in which models are learned to reason about the underlying physical dynamics of the particle systems of fluids from sequential visual observations.
We proposed \model{}, a differentiable deep learning framework that learns the 3D structures and physical dynamics of fluids by connecting particle-based dynamics modeling and particle-driven neural rendering.
The renderer learns to consider the geometric properties of fluid particles in volume rendering functions. 
It back-propagates the reconstruction errors between the synthesized and the observed images, thus allowing the transition model in \model{} to ground physical dynamics from visual observations.
\model{} was evaluated on tasks of fluid dynamics grounding, novel view synthesis, and future dynamics prediction.

\model{} still has some limitations that need to be further addressed in future work.
One is that it requires initial particle velocities. Further discussion is needed on how to estimate initial velocities or find a way to bypass this requirement.
Another issue is how to reduce the cumulative error of particle transitions, which may require defining effective physics-informed constraints in particle space.

\section*{Acknowledgements}
This work was supported by the Natural Science Foundation of China (U19B2035, 62106144), Shanghai Municipal Science and Technology Major Project (2021SHZDZX0102), and Shanghai Sailing Program (21Z510202133).

\bibliography{example_paper}

\begin{thebibliography}{51}
\providecommand{\natexlab}[1]{#1}
\providecommand{\url}[1]{\texttt{#1}}
\expandafter\ifx\csname urlstyle\endcsname\relax
  \providecommand{\doi}[1]{doi: #1}\else
  \providecommand{\doi}{doi: \begingroup \urlstyle{rm}\Url}\fi

\bibitem[Amada et~al.(2004)Amada, Imura, Yasumuro, Manabe, and
  Chihara]{amada2004particle}
Amada, T., Imura, M., Yasumuro, Y., Manabe, Y., and Chihara, K.
\newblock Particle-based fluid simulation on gpu.
\newblock In \emph{GPGPU}, volume~41, pp.\ ~42, 2004.

\bibitem[Battaglia et~al.(2013)Battaglia, Hamrick, and
  Tenenbaum]{battaglia2013simulation}
Battaglia, P.~W., Hamrick, J.~B., and Tenenbaum, J.~B.
\newblock Simulation as an engine of physical scene understanding.
\newblock \emph{Proceedings of the National Academy of Sciences}, 110\penalty0
  (45):\penalty0 18327--18332, 2013.

\bibitem[Bender \& Koschier(2015)Bender and Koschier]{bender2015divergence}
Bender, J. and Koschier, D.
\newblock Divergence-free smoothed particle hydrodynamics.
\newblock In \emph{SCA}, pp.\  147--155, 2015.

\bibitem[Biedert et~al.(2018)Biedert, Sohns, Schr{\"o}der, Amstutz, Wald, and
  Garth]{biedert2018direct}
Biedert, T., Sohns, J.-T., Schr{\"o}der, S., Amstutz, J., Wald, I., and Garth,
  C.
\newblock Direct raytracing of particle-based fluid surfaces using anisotropic
  kernels.
\newblock In \emph{EGPGV}, pp.\  1--12, 2018.

\bibitem[Chen et~al.(2021)Chen, Litalien, Gao, Wang, Fuji~Tsang, Khamis,
  Litany, and Fidler]{chen2021dib}
Chen, W., Litalien, J., Gao, J., Wang, Z., Fuji~Tsang, C., Khamis, S., Litany,
  O., and Fidler, S.
\newblock {DIB-R++}: Learning to predict lighting and material with a hybrid
  differentiable renderer.
\newblock In \emph{NeurIPS}, 2021.

\bibitem[Chentanez \& M{\"u}ller(2013)Chentanez and
  M{\"u}ller]{chentanez2013mass}
Chentanez, N. and M{\"u}ller, M.
\newblock Mass-conserving eulerian liquid simulation.
\newblock \emph{IEEE Transactions on Visualization and Computer Graphics},
  20\penalty0 (1):\penalty0 17--29, 2013.

\bibitem[Chorin(1968)]{chorin1968numerical}
Chorin, A.~J.
\newblock Numerical solution of the {Navier-Stokes} equations.
\newblock \emph{Mathematics of computation}, 22\penalty0 (104):\penalty0
  745--762, 1968.

\bibitem[Community(2018)]{blender}
Community, B.~O.
\newblock \emph{Blender - a {3D} modelling and rendering package}.
\newblock Blender Foundation, Stichting Blender Foundation, Amsterdam, 2018.
\newblock URL \url{http://www.blender.org}.

\bibitem[Fang et~al.(2020)Fang, Qu, Li, Zhang, Zhu, Aanjaneya, and
  Jiang]{fang2020iq}
Fang, Y., Qu, Z., Li, M., Zhang, X., Zhu, Y., Aanjaneya, M., and Jiang, C.
\newblock {IQ-MPM}: an interface quadrature material point method for
  non-sticky strongly two-way coupled nonlinear solids and fluids.
\newblock \emph{ACM Transactions on Graphics}, 39\penalty0 (4):\penalty0 51--1,
  2020.

\bibitem[Han et~al.(2020)Han, Chen, Liu, and Zwicker]{handrwr2020}
Han, Z., Chen, C., Liu, Y.-S., and Zwicker, M.
\newblock {DRWR}: A differentiable renderer without rendering for unsupervised
  3{D} structure learning from silhouette images.
\newblock In \emph{ICML}, 2020.

\bibitem[Holl et~al.(2020)Holl, Koltun, and Thuerey]{holl2020learning}
Holl, P., Koltun, V., and Thuerey, N.
\newblock Learning to control pdes with differentiable physics.
\newblock \emph{arXiv preprint arXiv:2001.07457}, 2020.

\bibitem[Hu et~al.(2019)Hu, Anderson, Li, Sun, Carr, Ragan-Kelley, and
  Durand]{hu2019difftaichi}
Hu, Y., Anderson, L., Li, T.-M., Sun, Q., Carr, N., Ragan-Kelley, J., and
  Durand, F.
\newblock {DiffTaichi}: Differentiable programming for physical simulation.
\newblock \emph{arXiv preprint arXiv:1910.00935}, 2019.

\bibitem[Jiang et~al.(2015)Jiang, Schroeder, Selle, Teran, and
  Stomakhin]{jiang2015affine}
Jiang, C., Schroeder, C., Selle, A., Teran, J., and Stomakhin, A.
\newblock The affine particle-in-cell method.
\newblock \emph{ACM Transactions on Graphics}, 34\penalty0 (4):\penalty0 1--10,
  2015.

\bibitem[Kajiya \& Von~Herzen(1984)Kajiya and Von~Herzen]{kajiya1984ray}
Kajiya, J.~T. and Von~Herzen, B.~P.
\newblock Ray tracing volume densities.
\newblock In \emph{SIGGRAPH}, volume~18, pp.\  165--174, 1984.

\bibitem[Kato et~al.(2018)Kato, Ushiku, and Harada]{kato2018neural}
Kato, H., Ushiku, Y., and Harada, T.
\newblock Neural {3D} mesh renderer.
\newblock In \emph{CVPR}, pp.\  3907--3916, 2018.

\bibitem[Kim et~al.(2019)Kim, Azevedo, Thuerey, Kim, Gross, and
  Solenthaler]{kim2019deep}
Kim, B., Azevedo, V.~C., Thuerey, N., Kim, T., Gross, M., and Solenthaler, B.
\newblock Deep fluids: A generative network for parameterized fluid
  simulations.
\newblock In \emph{Computer Graphics Forum}, volume~38, pp.\  59--70, 2019.

\bibitem[Kingma \& Ba(2015)Kingma and Ba]{kingma2015adam}
Kingma, D.~P. and Ba, J.
\newblock Adam: A method for stochastic optimization.
\newblock In \emph{ICLR}, 2015.

\bibitem[Ladick{\`y} et~al.(2015)Ladick{\`y}, Jeong, Solenthaler, Pollefeys,
  and Gross]{ladicky2015data}
Ladick{\`y}, L., Jeong, S., Solenthaler, B., Pollefeys, M., and Gross, M.
\newblock Data-driven fluid simulations using regression forests.
\newblock \emph{ACM Transactions on Graphics}, 34\penalty0 (6):\penalty0 1--9,
  2015.

\bibitem[Li et~al.(2018{\natexlab{a}})Li, Aittala, Durand, and
  Lehtinen]{li2018differentiable}
Li, T.-M., Aittala, M., Durand, F., and Lehtinen, J.
\newblock Differentiable monte carlo ray tracing through edge sampling.
\newblock \emph{ACM Transactions on Graphics}, 37\penalty0 (6):\penalty0 1--11,
  2018{\natexlab{a}}.

\bibitem[Li et~al.(2018{\natexlab{b}})Li, Wu, Tedrake, Tenenbaum, and
  Torralba]{li2018learning}
Li, Y., Wu, J., Tedrake, R., Tenenbaum, J.~B., and Torralba, A.
\newblock Learning particle dynamics for manipulating rigid bodies, deformable
  objects, and fluids.
\newblock In \emph{ICLR}, 2018{\natexlab{b}}.

\bibitem[Li et~al.(2019)Li, Wu, Tedrake, Tenenbaum, and
  Torralba]{li2019learning}
Li, Y., Wu, J., Tedrake, R., Tenenbaum, J.~B., and Torralba, A.
\newblock Learning particle dynamics for manipulating rigid bodies, deformable
  objects, and fluids.
\newblock In \emph{ICLR}, 2019.

\bibitem[Li et~al.(2020)Li, Lin, Yi, Bear, Yamins, Wu, Tenenbaum, and
  Torralba]{li2020visual}
Li, Y., Lin, T., Yi, K., Bear, D., Yamins, D., Wu, J., Tenenbaum, J., and
  Torralba, A.
\newblock Visual grounding of learned physical models.
\newblock In \emph{ICML}, pp.\  5927--5936, 2020.

\bibitem[Li et~al.(2022)Li, Li, Sitzmann, Agrawal, and Torralba]{li20223d}
Li, Y., Li, S., Sitzmann, V., Agrawal, P., and Torralba, A.
\newblock {3D} neural scene representations for visuomotor control.
\newblock In \emph{CoRL}, pp.\  112--123, 2022.

\bibitem[Li et~al.(2021)Li, Niklaus, Snavely, and Wang]{li2021neural}
Li, Z., Niklaus, S., Snavely, N., and Wang, O.
\newblock Neural scene flow fields for space-time view synthesis of dynamic
  scenes.
\newblock In \emph{CVPR}, 2021.

\bibitem[Liu et~al.(2021)Liu, Habermann, Rudnev, Sarkar, Gu, and
  Theobalt]{liu2021neural}
Liu, L., Habermann, M., Rudnev, V., Sarkar, K., Gu, J., and Theobalt, C.
\newblock Neural actor: Neural free-view synthesis of human actors with pose
  control.
\newblock \emph{ACM Transactions on Graphics}, 40\penalty0 (6):\penalty0 1--16,
  2021.

\bibitem[Liu et~al.(2019)Liu, Li, Chen, and Li]{liu2019soft}
Liu, S., Li, T., Chen, W., and Li, H.
\newblock Soft rasterizer: A differentiable renderer for image-based {3D}
  reasoning.
\newblock In \emph{ICCV}, pp.\  7708--7717, 2019.

\bibitem[Loper \& Black(2014)Loper and Black]{loper2014opendr}
Loper, M.~M. and Black, M.~J.
\newblock {OpenDR}: An approximate differentiable renderer.
\newblock In \emph{ECCV}, pp.\  154--169, 2014.

\bibitem[Mildenhall et~al.(2020)Mildenhall, Srinivasan, Tancik, Barron,
  Ramamoorthi, and Ng]{mildenhall2020nerf}
Mildenhall, B., Srinivasan, P.~P., Tancik, M., Barron, J.~T., Ramamoorthi, R.,
  and Ng, R.
\newblock {NeRF}: Representing scenes as neural radiance fields for view
  synthesis.
\newblock In \emph{ECCV}, pp.\  405--421, 2020.

\bibitem[Monaghan(1992)]{monaghan1992smoothed}
Monaghan, J.~J.
\newblock Smoothed particle hydrodynamics.
\newblock \emph{Annual Review of Astronomy and Astrophysics}, 30:\penalty0
  543--574, 1992.

\bibitem[Mrowca et~al.(2018)Mrowca, Zhuang, Wang, Haber, Fei-Fei, Tenenbaum,
  and Yamins]{mrowca2018flexible}
Mrowca, D., Zhuang, C., Wang, E., Haber, N., Fei-Fei, L., Tenenbaum, J.~B., and
  Yamins, D.~L.
\newblock Flexible neural representation for physics prediction.
\newblock In \emph{NeurIPS}, pp.\  8813--8824, 2018.

\bibitem[M{\"u}ller et~al.(2003)M{\"u}ller, Charypar, and
  Gross]{muller2003particle}
M{\"u}ller, M., Charypar, D., and Gross, M.~H.
\newblock Particle-based fluid simulation for interactive applications.
\newblock In \emph{SCA}, pp.\  154--159, 2003.

\bibitem[M{\"u}ller et~al.(1999)M{\"u}ller, Milano, and
  Koumoutsakos]{muller1999application}
M{\"u}ller, S., Milano, M., and Koumoutsakos, P.
\newblock Application of machine learning algorithms to flow modeling and
  optimization.
\newblock \emph{Annual Research Briefs}, pp.\  169--178, 1999.

\bibitem[Nguyen-Phuoc et~al.(2018)Nguyen-Phuoc, Li, Balaban, and
  Yang]{RenderNet2018}
Nguyen-Phuoc, T., Li, C., Balaban, S., and Yang, Y.-L.
\newblock {RenderNet}: A deep convolutional network for differentiable
  rendering from {3D} shapes.
\newblock In \emph{NeurIPS}, 2018.

\bibitem[Park et~al.(2021)Park, Sinha, Barron, Bouaziz, Goldman, Seitz, and
  Martin-Brualla]{park2021nerfies}
Park, K., Sinha, U., Barron, J.~T., Bouaziz, S., Goldman, D.~B., Seitz, S.~M.,
  and Martin-Brualla, R.
\newblock Nerfies: Deformable neural radiance fields.
\newblock In \emph{ICCV}, pp.\  5865--5874, 2021.

\bibitem[Peng et~al.(2021)Peng, Dong, Wang, Zhang, Shuai, Zhou, and
  Bao]{peng2021animatable}
Peng, S., Dong, J., Wang, Q., Zhang, S., Shuai, Q., Zhou, X., and Bao, H.
\newblock Animatable neural radiance fields for modeling dynamic human bodies.
\newblock In \emph{ICCV}, pp.\  14314--14323, 2021.

\bibitem[Poloni et~al.(2000)Poloni, Giurgevich, Onesti, and
  Pediroda]{poloni2000hybridization}
Poloni, C., Giurgevich, A., Onesti, L., and Pediroda, V.
\newblock Hybridization of a multi-objective genetic algorithm, a neural
  network and a classical optimizer for a complex design problem in fluid
  dynamics.
\newblock \emph{Computer Methods in Applied Mechanics and Engineering},
  186\penalty0 (2-4):\penalty0 403--420, 2000.

\bibitem[Price(2012)]{price2012smoothed}
Price, D.~J.
\newblock Smoothed particle hydrodynamics and magnetohydrodynamics.
\newblock \emph{Journal of Computational Physics}, 231\penalty0 (3):\penalty0
  759--794, 2012.

\bibitem[Pumarola et~al.(2021)Pumarola, Corona, Pons-Moll, and
  Moreno-Noguer]{pumarola2021d}
Pumarola, A., Corona, E., Pons-Moll, G., and Moreno-Noguer, F.
\newblock {D-NeRF}: Neural radiance fields for dynamic scenes.
\newblock In \emph{CVPR}, pp.\  10318--10327, 2021.

\bibitem[Qi et~al.(2017)Qi, Yi, Su, and Guibas]{qi2017pointnetpp}
Qi, C.~R., Yi, L., Su, H., and Guibas, L.~J.
\newblock Point{N}et++: Deep hierarchical feature learning on point sets in a
  metric space.
\newblock In \emph{NeurIPS}, 2017.

\bibitem[Sanchez-Gonzalez et~al.(2020)Sanchez-Gonzalez, Godwin, Pfaff, Ying,
  Leskovec, and Battaglia]{sanchez2020learning}
Sanchez-Gonzalez, A., Godwin, J., Pfaff, T., Ying, R., Leskovec, J., and
  Battaglia, P.
\newblock Learning to simulate complex physics with graph networks.
\newblock In \emph{ICML}, pp.\  8459--8468, 2020.

\bibitem[Schenck \& Fox(2018)Schenck and Fox]{schenck2018spnets}
Schenck, C. and Fox, D.
\newblock {SPNets}: Differentiable fluid dynamics for deep neural networks.
\newblock In \emph{CoRL}, pp.\  317--335, 2018.

\bibitem[Stam(1999)]{stam1999stable}
Stam, J.
\newblock Stable fluids.
\newblock In \emph{SIGGRAPH}, pp.\  121--128, 1999.

\bibitem[Stomakhin et~al.(2013)Stomakhin, Schroeder, Chai, Teran, and
  Selle]{stomakhin2013material}
Stomakhin, A., Schroeder, C., Chai, L., Teran, J., and Selle, A.
\newblock A material point method for snow simulation.
\newblock \emph{ACM Transactions on Graphics}, 32\penalty0 (4):\penalty0 1--10,
  2013.

\bibitem[Su et~al.(2021)Su, Yu, Zollh{\"o}fer, and Rhodin]{su2021nerf}
Su, S.-Y., Yu, F., Zollh{\"o}fer, M., and Rhodin, H.
\newblock {A-NeRF}: Articulated neural radiance fields for learning human
  shape, appearance, and pose.
\newblock In \emph{NeurIPS}, 2021.

\bibitem[Sulsky et~al.(1995)Sulsky, Zhou, and Schreyer]{sulsky1995application}
Sulsky, D., Zhou, S.-J., and Schreyer, H.~L.
\newblock Application of a particle-in-cell method to solid mechanics.
\newblock \emph{Computer Physics Communications}, 87\penalty0 (1-2):\penalty0
  236--252, 1995.

\bibitem[Ummenhofer et~al.(2019)Ummenhofer, Prantl, Thuerey, and
  Koltun]{ummenhofer2019lagrangian}
Ummenhofer, B., Prantl, L., Thuerey, N., and Koltun, V.
\newblock Lagrangian fluid simulation with continuous convolutions.
\newblock In \emph{ICLR}, 2019.

\bibitem[Wang et~al.(2004)Wang, Bovik, Sheikh, and Simoncelli]{wang2004image}
Wang, Z., Bovik, A.~C., Sheikh, H.~R., and Simoncelli, E.~P.
\newblock Image quality assessment: from error visibility to structural
  similarity.
\newblock \emph{IEEE Transactions on Image Processing}, 13\penalty0
  (4):\penalty0 600--612, 2004.

\bibitem[Wu et~al.(2017{\natexlab{a}})Wu, Lu, Kohli, Freeman, and
  Tenenbaum]{wu2017learning}
Wu, J., Lu, E., Kohli, P., Freeman, B., and Tenenbaum, J.
\newblock Learning to see physics via visual de-animation.
\newblock In \emph{NeurIPS}, volume~30, pp.\  153--164, 2017{\natexlab{a}}.

\bibitem[Wu et~al.(2017{\natexlab{b}})Wu, Wang, Xue, Sun, Freeman, and
  Tenenbaum]{marrnet}
Wu, J., Wang, Y., Xue, T., Sun, X., Freeman, W.~T., and Tenenbaum, J.~B.
\newblock {MarrNet}: {3D} shape reconstruction via {2.5D} sketches.
\newblock In \emph{NeurIPS}, 2017{\natexlab{b}}.

\bibitem[Zhang et~al.(2018)Zhang, Isola, Efros, Shechtman, and
  Wang]{zhang2018unreasonable}
Zhang, R., Isola, P., Efros, A.~A., Shechtman, E., and Wang, O.
\newblock The unreasonable effectiveness of deep features as a perceptual
  metric.
\newblock In \emph{CVPR}, pp.\  586--595, 2018.

\bibitem[Zhu \& Bridson(2005)Zhu and Bridson]{zhu2005animating}
Zhu, Y. and Bridson, R.
\newblock Animating sand as a fluid.
\newblock \emph{ACM Transactions on Graphics}, 24\penalty0 (3):\penalty0
  965--972, 2005.

\end{thebibliography}
\bibliographystyle{icml2022}

\end{document}